\documentclass[sigconf]{acmart}

\usepackage{multicol}
\usepackage{multirow}
\usepackage{amsmath, amsfonts, amsthm}
\usepackage{bm}
\usepackage{tabularx}
\usepackage{graphicx}
\usepackage[linesnumbered,ruled,vlined]{algorithm2e}
\usepackage{caption}
\usepackage[list=true]{subcaption}
\usepackage{bbm}
\usepackage{float}
\usepackage{colortbl}
\usepackage{balance}
\usepackage{booktabs}
\usepackage{tcolorbox}
\usepackage{enumitem}
\usepackage{arydshln} 
\usepackage{makecell}

\setlistdepth{9}
\setlist[itemize,1]{label=$\bullet$}
\setlist[itemize,2]{label=$\bullet$}
\setlist[itemize,3]{label=$\circ$}
\setlist[itemize,4]{label=$\ast$}
\setlist[itemize,5]{label=$\diamond$}
\setlist[itemize,6]{label=$\bullet$}
\setlist[itemize,7]{label=$\bullet$}
\setlist[itemize,8]{label=$\bullet$}
\setlist[itemize,9]{label=$\bullet$}
\renewlist{itemize}{itemize}{9}

\copyrightyear{2026}
\acmYear{2026}
\setcopyright{cc}
\setcctype{by}
\acmConference[MM '26] {Proceedings of the 34th ACM International Conference on Multimedia}{November 10--14, 2026}{Rio de Janeiro, Brazil.}
\acmBooktitle{Proceedings of the 34th ACM International Conference on Multimedia (MM '26), November 10--14, 2026, Rio de Janeiro, Brazil}
\acmISBN{979-8-4007-2213-4/2026/11}
\acmDOI{10.1145/3767308.3836334}

\newcommand{\hide}[1]{} 

\title{From Cheap Fakes to Pure Synthesis: Addressing the New Era of T2V Fake News Videos}
\thanks{Our code can be found in \url{https://github.com/TrustworthyComp/PS-FNVD}.}

\author{Yifeng Luo}
\affiliation{
    \department{Department of Interactive Media\\Hong Kong Baptist University}
    \department{AI and Social Good Lab, AI Media Centre\\Hong Kong Baptist University}
    \city{Hong Kong}
    \country{China}
}
\email{yfengL@life.hkbu.edu.hk}
\orcid{0000-0003-3635-8154}

\author{Yupeng Li}
\authornote{This work was done while Yifeng Luo was under the supervision of Yupeng Li and Liang Lan. Yupeng Li is the corresponding author.}
\affiliation{
    \department{Department of Interactive Media\\Hong Kong Baptist University}
    \department{AI and Social Good Lab, AI Media Centre\\Hong Kong Baptist University}
    \city{Hong Kong}
    \country{China}
}
\email{ivanypli@gmail.com}

\author{Liang Lan}
\affiliation{
    \department{Department of Interactive Media\\Hong Kong Baptist University}
    \department{AI Media Centre\\Hong Kong Baptist University}
    \city{Hong Kong}
    \country{China}
}
\email{lanliang@hkbu.edu.hk}

\author{Tian Wang}
\affiliation{
    \department{Institute of Artificial Intelligence and Future Networks}
    \institution{Beijing Normal University}
    \city{Zhuhai}
    \country{China}
}
\email{tianwang@bnu.edu.cn}

\begin{document}

\begin{abstract}
Recent text-to-video (T2V) generation models enable fake news videos to be synthesized from scratch, shifting the threat beyond cheap fakes assembled from existing footage. 
Such news videos can closely match fabricated narratives, creating a modality alignment trap for existing detectors.
Existing datasets lack pure synthesis fake news videos. 
Although directly prompting T2V models with descriptions of fake news videos can yield perfectly aligned samples, it reduces the fake news video detection (FNVD) to unimodal shortcuts and causes semantic-visual degeneration.
To counter this, we formulate T2V-FNVD as a novel ternary classification task with three labels (real, cheap fake, and pure synthesis fake) and construct the first pure synthesis fake news video dataset (PS-FNVD). 
PS-FNVD includes fabricated events with aligned deception (Type 1) and true events with false visual provenance (Type 2), preventing models from exploiting unimodal shortcuts.
Furthermore, we propose the Reasoning-guided T2V-FNVD (R-T2V) framework. Trained through conditioned rationale generation and supervised fine-tuning, R-T2V integrates high-level semantic logic with low-level physical generative traces to predict the ternary veracity label.
Extensive experiments across 10 prevailing baselines show that R-T2V achieves the state-of-the-art performance, outperforming the second-best baseline by 12.20 percentage points in accuracy and 8.46 percentage points in macro $F_1$.
\end{abstract}



\maketitle
\section{Introduction}
\label{Sec:Intr}

Short-video platforms have become an important channel for daily news consumption, and the prevalence of fake-news videos has increased alongside their growing use.
Existing studies define a fake news video as a video post that conveys false, inaccurate, or misleading information \cite{bu2024fakingrecipe, qi2023fakesv, bu2023combating, bu2025enhancing}.
Such fake news videos spread faster than textual or image-based misinformation of the same content and are perceived as more trustworthy by users, posing substantial risks to public opinion and social good \cite{sundar2021seeing, he2026fact2fiction, he2026debating, luo2024message}.

\begin{figure}[t]
\centering 
\includegraphics[scale=0.98]{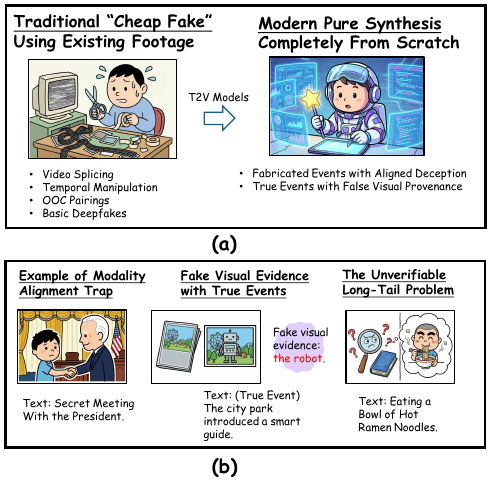}
\caption{(a) An illustration of traditional \textit{cheap fakes} and \textit{modern pure synthesis fakes}. (b) An example of \textit{modality alignment trap}, an example of fake visual evidence with true event, and the \textit{unverifiable long-tail problem}.}
\label{cfvsps}
\end{figure}

Traditionally, the creation of fake news videos was constrained by the availability of source material. Fabricators relied on a spectrum of \textit{cheap fakes} \cite{paris2019deepfakes}, such as video splicing, temporal manipulation, contextually deceptive out-of-context (OOC) pairings, and basic deepfakes like face-swaps \cite{bu2024fakingrecipe, bu2025enhancing, bu2023combating, shen2024comprehensive, qi2024sniffer, papadopoulos2025similarity}. Because these methods force existing disparate footage to fit a fabricated narrative, they inherently leave visible editing traces or semantic compromises \cite{bu2023combating}.
Recently, text-to-video (T2V) generation models have achieved remarkable advancements in the field of video generation \cite{han2025video, brooks2024video}. The state-of-the-art (SOTA) T2V generation models have fundamentally shifted the fake news video threat from \textit{cheap fakes} to \textit{pure synthesis}. Fabricators are no longer bound by reality; instead of painstakingly searching for and splicing old videos, fabricators can now simply prompt the T2V model to synthesize highly realistic, non-existent events from scratch (see Fig.~\ref{cfvsps}).
Critically, this threat cannot be mitigated by deploying AI-generated video detectors alone because Fake News Video Detection (FNVD) also requires 
understanding the interaction between the deceptive text and the synthesized video, rather than only identifying synthetic pixels.

Recent studies on FNVD and AI-generated Video Detection (AVD) have rapidly evolved, yet struggle against the threat of pure synthesis fake news videos. 
On the data front, existing datasets fall into two trajectories. First, large-scale FNVD datasets \cite{wang2024official, wang2025fmnv, zhang2025fact, qi2023fakesv, bu2024fakingrecipe} have introduced Large Language Models (LLMs) augmentation and fine-grained annotations; however, they still rely on modifying text entities or splicing existing real-world footage. Consequently, they essentially create OOC cheap fakes \cite{shen2024comprehensive, qi2024sniffer, papadopoulos2025similarity} bound by existing footage. Detecting OOC cheap fakes and countering pure synthesis fake news videos generated from scratch are two distinct problems. Second, other large-scale AVD datasets \cite{ni2026genvidbench, leotta2026synthforensics, veeramachaneni2025leveraging} focus on pure synthesis but are designed solely for AVD. They lack the necessary fake news context. 
On the method front, FNVD methods fall into two sub-streams. The first stream relies on traditional neural networks for multimodal representation fusion \cite{bu2024fakingrecipe, qi2023fakesv, shang2021multimodal, choi2021using, liu2023covid, qi2023two}. These models are primarily designed to detect cheap fakes by identifying cross-modal inconsistencies, such as OOC pairings between text and video. However, modern T2V models introduce the \textit{modality alignment trap} by generating videos that perfectly align with the fabricated textual narrative, making traditional cross-modal inconsistency checks useless. The second stream employs LLMs for data augmentation and reasoning with/or external evidence retrieved from search engines \cite{hong2025following, niu2025pioneering}, or directly fine-tunes LLMs on annotated text-video pairs \cite{zhang2025fact}. These approaches share a hidden assumption: \textit{they treat the video as a semantic container and rely on internal knowledge (resp.~external knowledge) from LLMs (resp.~search engines) to output a veracity prediction}. This assumption collapses when faced with pure synthesis fake news videos. First, these methods are vulnerable to fake visual evidence attached to true events. Second, these methods will fail due to the \textit{unverifiable long-tail problem}, because with the help of T2V models, fabricators can create highly personalized, unrecorded, or emerging events. Search engines will find no records and output \texttt{not enough evidence} (see Fig.~\ref{cfvsps}).

\begin{figure*}[t]
\centering 
\includegraphics[scale=0.85]{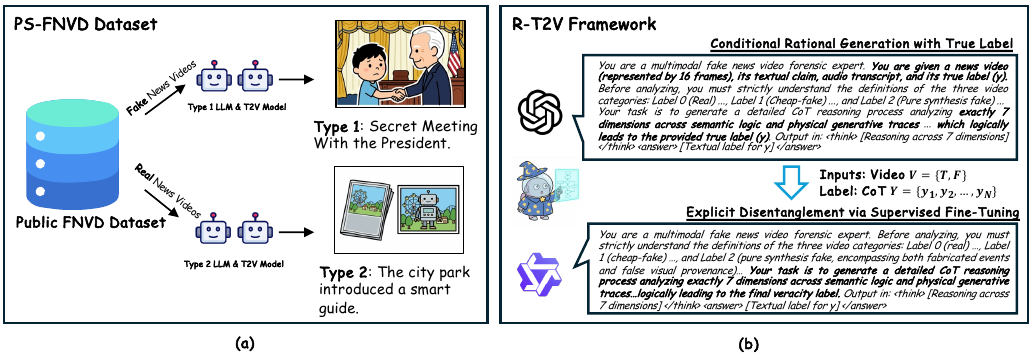}
\caption{(a) The overview of PS-FNVD dataset construction. Type 1: Fabricated events with aligned deception. Type 2: True events with false visual provenance. (b) The overview of the R-T2V framework.}
\label{dataset_overview}
\end{figure*}

The community lacks an FNVD dataset that contains pure synthesis fake news videos that jointly capture generative deception and complex semantic manipulation, as well as a method to counter it. 
In this study, we formulate a novel task tailored for the new era of fake news video, \textit{T2V-era Fake News Video Detection} (T2V-FNVD). Unlike existing FNVD studies that formulate the problem as a binary classification, i.e., real or fake, we define T2V-FNVD as a ternary classification problem. Given a news video, the model must classify it as real, cheap fake, or pure synthesis fake, while simultaneously generating high-quality explanations to justify its veracity prediction.
It is challenging to address the T2V-FNVD problem.
First, constructing a T2V-FNVD dataset is challenging.
Directly prompting the T2V models with the descriptions of fake news videos yields perfectly aligned synthetic fake news videos. Although these perfectly aligned samples successfully expose the vulnerabilities of existing FNVD methods, this generation strategy introduces the \textit{semantic-visual degeneration} problem. 
The task can then be solved trivially through unimodal shortcuts (i.e., pure text-based fact-checking or vision-based artifact detection) \cite{goyal2017making}, failing to equip models with the complex cross-modal reasoning required for the T2V-FNVD task.
Second, unifying the detection of cheap fakes and pure synthesis within a framework is difficult. Existing AVD methods target visual artifacts, ignoring the semantic context of news videos. Conversely, traditional FNVD methods rely on cross-modal inconsistencies or internal/external knowledge retrieval, both of which fail against the modality alignment trap, fake visual evidence with true events, and the unverifiable long-tail nature of T2V generation. Therefore, the core challenge lies in designing a framework that simultaneously evaluates semantic logic and physical generative traces to classify the veracity of news videos.

To address the challenges of the T2V-FNVD task, we introduce the first \textit{P}ure \textit{S}ynthesis \textit{FNVD} dataset, PS-FNVD, building upon the public FNVD dataset and a SOTA T2V generation model (Hunyuan Video \cite{kong2024hunyuanvideo}), alongside R-T2V, a novel \textit{R}easoning-guided \textit{T2V}-FNVD framework.
To construct the PS-FNVD dataset, we systematically model the generative threat by formulating two variants of T2V fake news videos: Fabricated Events with Aligned Deception (Type 1), which explicitly create the modality alignment trap, and True Events with False Visual Provenance (Type 2), which tackle the semantic-visual degeneration problem. This controlled generation pipeline yields inherently paired videos that share similar semantic contexts but differ in veracity and visual provenance, laying a crucial foundation for the T2V-FNVD model training.  
To address the joint reasoning challenge, we propose the R-T2V framework. The core of R-T2V is a data-centric reasoning topology that evaluates videos across seven predefined dimensions, directly bridging semantic logic and physical generative traces. To train this topology, we employ conditioned rationale generation. Specifically, we use ground-truth veracity labels to constrain a commercial Multimodal LLM (MLLM), generating Chain-of-Thought (CoT) trajectories. 
This strategy anchors each rationale to the correct label, preventing the teacher model's own misclassifications from entering the training data. Finally, through Supervised Fine-Tuning (SFT) on structured reasoning trajectories, R-T2V explicitly evaluates semantic logic and physical generative traces before outputting the final label, enabling the disentanglement of cheap fakes from pure synthesis fakes. The main contributions are summarized as follows:
\begin{itemize}
    \item \textbf{Novel Task and Dataset:} To the best of our knowledge, we are the first to study the problem of T2V-FNVD, formulating it as a novel ternary classification task tailored for the generative AI era. To support this, we construct PS-FNVD, the first pure synthesis fake news video dataset. By systematically generating Type 1 and Type 2 videos, the dataset explicitly models the modality alignment trap and semantic-visual degeneration, providing inherently paired data that prevents unimodal shortcuts.

    \item \textbf{Reasoning-guided Detection Framework:} We propose R-T2V, a reasoning-guided framework built on a seven reasoning dimensions topology that bridges semantic logic and physical generative traces.
    Through conditioned rationale generation and SFT, R-T2V learns to reason across these dimensions for ternary classification.

    \item \textbf{SOTA Performance:} Extensive experiments across zero-shot MLLMs, agent-based reasoning, and supervised training paradigms show that R-T2V achieves SOTA performance, exceeding the second-best
    baseline by 12.20 (resp.~8.46) percentage points in accuracy (resp.~macro $F_1$).
\end{itemize}

\section{T2V Fake News Video Detection}

\subsection{Problem Setting}
Let $V = \{T, F\}$ denote a news video, where $T$ is the news video corresponding text and audio transcript, and $F$ is the video frames. 
The task of T2V-FNVD is formulated as a ternary problem. Given an input news video $V$, the goal is to learn a fake news video detector $f_\theta: \{T, F\} \rightarrow \hat{y} \times E$, where $\hat{y} \in \{0, 1, 2\}$ is the predicted class
indicating whether the video is 
real ($y=0$), cheap-fake ($y=1$), or pure synthesis fake ($y=2$); and $E$ is the explanation.
Our setting addresses the threat of generative AI. In our dataset $\mathcal{D} = \{(V_i, y_i)\}_{i=1}^N$, for a fabricated news video $V_\text{fake} = \{T_\text{fake}, F_\text{fake}\}$, the video frames $F_\text{fake}$ are synthesized by a T2V generation model $\mathcal{G}$ conditioned on the fabricated context (i.e., $F_\text{fake} \sim \mathcal{G}(p)$). The fake news video detector $f_\theta(\cdot)$ is trained on $\mathcal{D}_\text{train}$ and evaluated on $\mathcal{D}_\text{test}$, both of which contain these pure synthesis fake news videos.

\subsection{Pure Synthesis FNVD Dataset Construction}
To simulate the threat of modern generative deception, we construct the PS-FNVD dataset by leveraging an existing open-source FNVD dataset (FakeSV \cite{qi2023fakesv}) and SOTA T2V generation model (i.e., Hunyuan Video \cite{kong2024hunyuanvideo}) as our foundation. The base dataset contains both authentic and traditional cheap fake news videos.
To systematically generate the variants of pure synthesis fake news videos defined in our taxonomy, we introduce two specialized LLMs acting as prompt writers, denoted as $\mathcal{P}_1$ and $\mathcal{P}_2$. Their objective is to generate highly descriptive T2V prompts that maximize the generative capabilities of the modern T2V generator $\mathcal{G}$ while fulfilling specific deceptive intents. 
An additional critical objective of these writers is to navigate the strict safety alignment mechanisms of commercial models \cite{ji2023beavertails, lu2025alignment}. Directly prompting with sensitive and disaster-related fake news content often triggers these filters, resulting in explicit safety refusals from the models. Therefore, we strategically employ bypass techniques, such as context manipulation \cite{zheng2026riskatlas}, objective visual focus, and moderated entity generalization \cite{ba2024surrogateprompt}. This enables high generation success rates for defensive research. Detailed prompts and comprehensive bypass strategies are provided in Appendix~\ref{APP:Dataset}.
The overview of the PS-FNVD dataset construction is presented in Fig.~\ref{dataset_overview}.

\paragraph{Type 1: Fabricated Events with Aligned Deception} Our goal is to create the modality alignment trap. We utilize the traditional cheap fake news videos $V_\text{fake}^\text{cheap}$ from the base dataset. The Type 1 prompt writer $\mathcal{P}_1$ takes $V_\text{fake}^\text{cheap}$ as input and expands it into a detailed, visually rich prompt $p_1 = \mathcal{P}_1(V_\text{fake}^\text{cheap})$ that perfectly aligns with the deceptive narrative. The T2V generator $\mathcal{G}$ then generates the synthetic frames $F^{T1}_\text{fake} \sim \mathcal{G}(p_1)$. The resulting Type 1 fake news video is formulated as $V^{T1}_\text{fake} = \{T_\text{fake}, F^{T1}_\text{fake}\}$. 

\paragraph{Type 2: True Events with False Visual Provenance} We simulate the scenario where fabricators attach fake visual evidence to a real event. We utilize the authentic real news videos $V_\text{real}$ from the base dataset, retaining their corresponding text. The Type 2 prompt writer $\mathcal{P}_2$ takes $V_\text{real}$ and is instructed to conceptualize an exaggerated, sensationalized, or contextually displaced visual scene, yielding the prompt $p_2 = \mathcal{P}_2(V_\text{real})$. For instance, a minor earthquake report might be paired with a prompt describing a catastrophic tsunami. The T2V generator $\mathcal{G}$ then synthesizes these frames $F_\text{fake}^{T2} \sim \mathcal{G}(p_2)$. The resulting Type 2 fake news video is $V^{T2}_\text{fake} = \{T_\text{real}, F_\text{fake}^{T2}\}$. Here, the semantic premise is verifiable and true, but the visual provenance is entirely fabricated. Through this controlled generation pipeline, our dataset inherently contains paired videos that share similar semantic contexts but differ in veracity and visual provenance.

\paragraph{Pure Synthesis Fake News Video Detection Dataset Statistics}
The PS-FNVD dataset comprises 6,636 news videos, structured to support the ternary classification task. The open-source FNVD dataset (FakeSV \cite{qi2023fakesv}) contains 1,687 real news videos (Label 0), alongside 1,631 cheap fakes (Label 1) constructed from existing footage.\footnote{The original FakeSV contains 1,827 fake and 1,827 real news videos. We filtered out videos that are no longer downloadable.} To represent pure synthesis videos (Label 2), we synthesized a total of 3,318 videos using a strict one-to-one mapping strategy. Specifically, we generated 1,631 Type 1 videos based on the cheap fake narratives, and 1,687 Type 2 videos derived from the real news textual claims. This inherently paired distribution prevents the target model from exploiting text-only class-prior biases, providing a robust data foundation for training the explicit disentanglement.\footnote{Throughout, we use explicit disentanglement in a behavioral rather than representational sense: the model is required to articulate a separate verdict for semantic logic-level and physical generative-level evidence before committing to a label. We make no claim about the structure of its internal representations.}
Table~\ref{tab:statistics} presents the statistics of our PS-FNVD dataset.

\begin{table}[!t]
\caption{The statistics of the PS-FNVD dataset.}
\label{tab:statistics}
\centering
\small
\begin{tabular}{cc}
\toprule
Label & Total Videos \\
\midrule
Real & 1687\\
Cheap Fake & 1631 \\
Type 1 & 1631 \\
Type 2 & 1687 \\
\bottomrule
\end{tabular}
\end{table}

\begin{figure}[!t]
\centering
\includegraphics[scale=0.41]{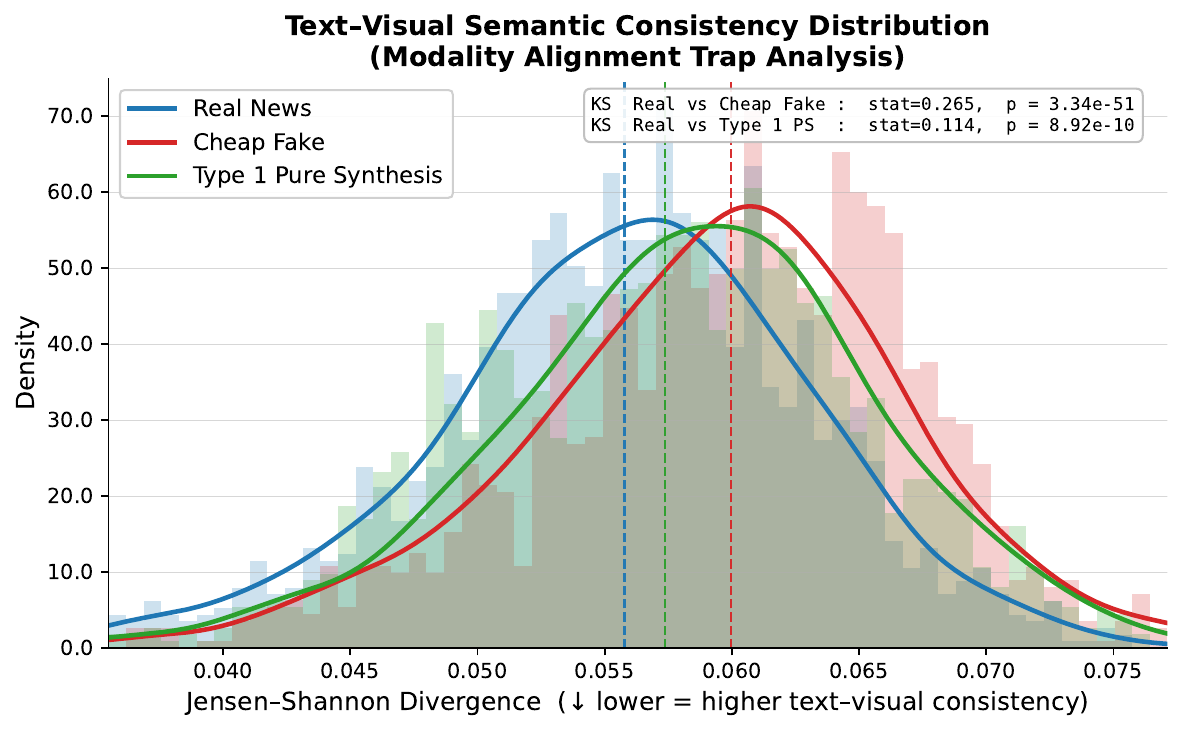}
\caption{\textbf{Distribution of text-visual semantic consistency across video categories.} We compute Jensen-Shannon divergence between CLIP text and frame embeddings (lower values indicate higher consistency). Cheap Fakes constructed from existing footage (red) exhibit substantially weaker alignment (mean JS $0.060$) than Real News (blue, mean $0.056$). Conversely, Type 1 Pure Synthesis generated completely from scratch (green) achieves alignment much closer to the authentic baseline (mean $0.057$). Two-sample Kolmogorov-Smirnov (KS) test results (inset) confirm that the distributional gap for Cheap Fakes ($D=0.265$, $p<0.001$) is over twice the magnitude of Type 1 Pure Synthesis ($D=0.114$, $p<0.001$), empirically validating the Modality Alignment Trap.}
\label{fig:alignment}
\end{figure}

\paragraph{Quantitative Alignment Assessment}
To empirically validate the existence of the modality alignment trap, we quantify text-visual semantic consistency by computing the Jensen-Shannon (JS) divergence between CLIP \cite{radford2021learning} text and video frame embeddings. As illustrated in Fig.~\ref{fig:alignment}, cheap fakes constructed from existing footage exhibit a higher mean JS divergence ($0.060$), reflecting the inherent cross-modal inconsistencies caused by forced narrative pairings. Conversely, Type 1 pure synthesis generated completely from scratch achieves a mean divergence of $0.057$, closely tracking the real news video baseline ($0.056$). 
This 
small difference is an expected artifact of the moderated entity generalization required to bypass the safety alignment of both the prompt writer LLM and the T2V generator, which
slightly bounds the visual extremity of the fabricated event to comply with model moderation.
A two-sample Kolmogorov-Smirnov (KS) test confirms a substantial distributional gap between real news and cheap fakes ($D=0.265$, $p < 0.001$). In contrast, the distributional shift between real news and Type 1 pure synthesis is less than half that magnitude ($D=0.114$, $p < 0.001$). 
Both gaps remain statistically detectable at this sample size; what differs is their magnitude.
This empirical evidence confirms that Type 1 pure synthesis successfully mimics the semantic alignment of real news, invalidating traditional FNVD methods that rely on detecting cross-modal inconsistencies.

\subsection{Reasoning-guided Fake News Video Detection Framework}

To address the T2V-FNVD task, we propose the R-T2V framework. Instead of mapping multimodal inputs directly to discrete labels, R-T2V enforces explicit verdict prediction through a data-centric reasoning topology. As outlined in Table \ref{tab:7_dimensions}, this topology evaluates videos across semantic logic and physical generative traces, enabling the model to distinguish cheap fakes constructed from existing footage from pure synthesis generated completely from scratch. The overview of the R-T2V framework is presented in Fig.~\ref{dataset_overview}. The detailed system prompt can be found in Appendix~\ref{APP:Dataset}.

\paragraph{Conditioned Rationale Generation}
To operationalize this topology, we construct a structured CoT reasoning corpus. For each video $V = \{T, F\}$ in our dataset, we extract the corresponding text and audio transcript $T$, and uniformly sample $16$ frames $F$. We employ conditioned rationale generation by using the ground-truth veracity label $y \in \{0, 1, 2\}$ to constrain a commercial MLLM (i.e., GPT-4o \cite{hurst2024gpt}). The MLLM is prompted to generate a reasoning trajectory across the seven predefined dimensions.
This explicit conditioning enables the generated rationales to terminate at the correct label rather than the teacher model's own prediction, yielding rationales with correct verdicts and dimension-level evidence.

\begin{table}[!t]
\caption{The seven dimensions for fake news video reasoning.}
\label{tab:7_dimensions}
\centering
\small
\begin{tabular}{p{0.15\linewidth} p{0.5\linewidth}}
\toprule
\textbf{Category} & \textbf{Dimension} \\
\midrule
\multirow{3}{*}{\shortstack{FNVD\\(Semantic \\\& Logical)}} 
& Text-Video Semantic Consistency \\
& Event Logical Plausibility \\
& Sensationalism \& Intent  \\
\midrule
\multirow{4}{*}{\shortstack{AVD\\(Visual \\\& Physical)}} 
& Spatial \& Physics Violations  \\
& Temporal Inconsistency \\
& Texture \& Lighting Anomalies \\
& Biological/Human Artifacts \\
\bottomrule
\end{tabular}
\end{table}

\begin{figure}[!t]
\centering
\includegraphics[scale=0.31]{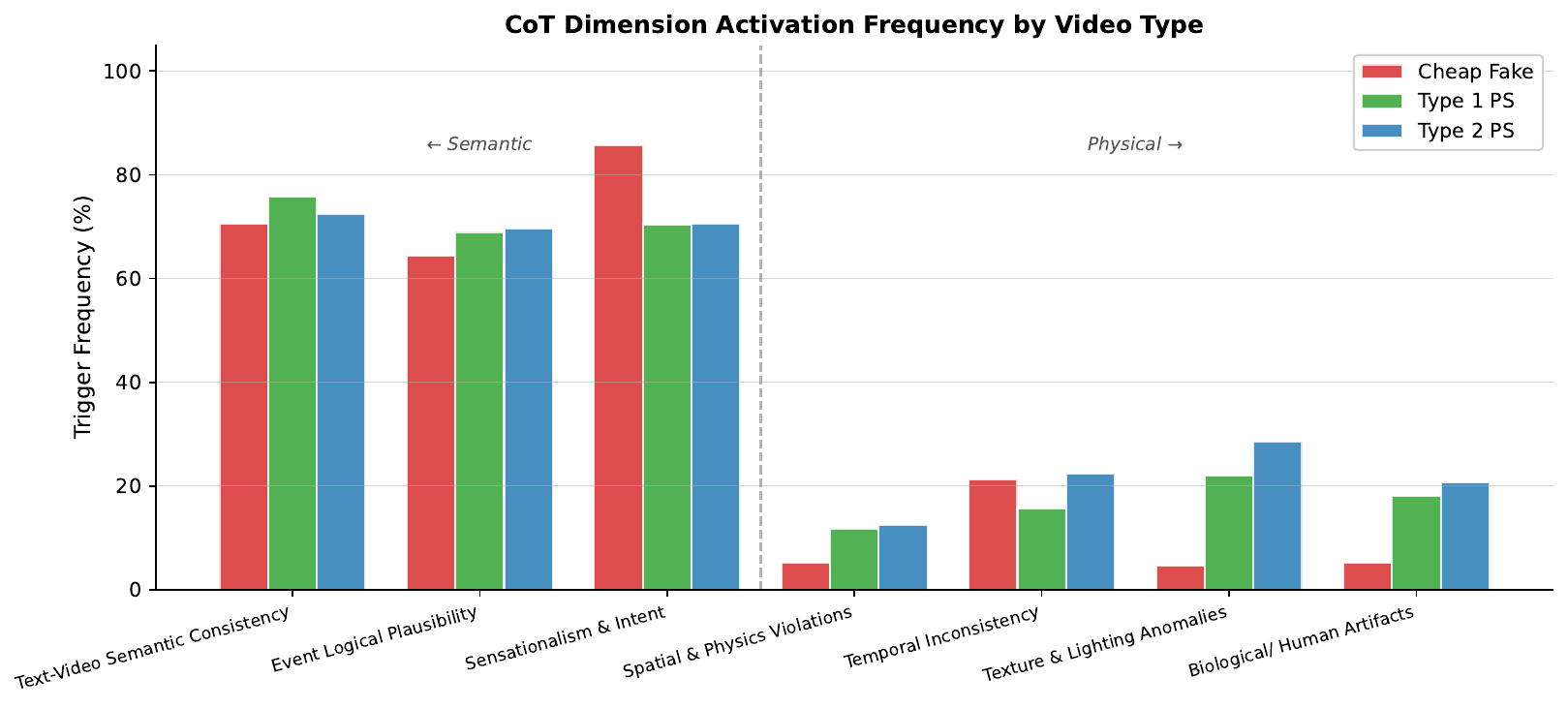}
\caption{\textbf{Activation frequency of reasoning dimensions in the generated CoT corpus.} The dashed line explicitly separates semantic logic (left) from physical generative traces (right). Cheap fakes constructed from existing footage (red) heavily trigger semantic evaluations while showing lower activation across most physical dimensions.
Conversely, pure synthesis generated completely from scratch (green and blue) exhibits higher activation rates for physical generative traces, validating the explicit disentanglement within our training data.}
\label{fig:cot_activation}
\end{figure}

\paragraph{CoT Reasoning Corpus Disentanglement Analysis}
To assess whether the generated rationales are informative rather than formulaic, we analyze the activation frequencies of the seven predefined dimensions within the generated CoT trajectories. 
A dimension is activated when the rationale flags a concrete anomaly rather than clearing it (e.g., \texttt{no visible physics violations}).
As illustrated in Fig.~\ref{fig:cot_activation}, the activation frequencies differ across the three fabrication types, rather than following a single pattern shared by all classes.
For cheap fakes constructed from existing footage, the trajectories predominantly trigger semantic logic dimensions, peaking at $85.7\%$ for Sensationalism \& Intent. 
Concurrently, they exhibit negligible activation (under $6\%$) across most physical generative traces (excluding temporal inconsistency), consistent with these videos being predominantly assembled from camera-captured footage. The residual activations correspond to a small number of hybrid cases in which AI-generated segments are spliced into otherwise authentic material, a mixture that neither the cheap-fake nor the pure-synthesis category fully captures.
In contrast, trajectories for pure synthesis generated completely from scratch demonstrate a substantial spike in physical evaluations, activating Texture \& Lighting Anomalies and Biological/Human Artifacts $4\times$ to $6\times$ more frequently compared to cheap fakes. 
Notably, while both Type 1 and Type 2 pure synthesis exhibit comparable semantic activations, confirming their shared capability in high-level contextual manipulation, 
Type 2 videos exhibit modestly higher physical-anomaly rates (e.g., $28.6\%$ vs. $22.1\%$ for texture anomalies), consistent with the hypothesis that prompting for contextually displaced or exaggerated scenes pushes the generator further from its training distribution.
Furthermore, as our PS-FNVD dataset generation incorporates moderated exaggeration to bypass safety filters, the synthesized videos avoid extreme shock value (e.g., explicit gore). 
Ultimately, this stark divergence indicates that our generated corpus successfully achieves explicit disentanglement, forcing the target model to separate contextual manipulation from physical anomalies during training.

\paragraph{Explicit Disentanglement via Supervised Fine-Tuning}
Using these structured CoT trajectories, we apply SFT to a base MLLM (Qwen2.5-VL-7B \cite{bai1others}), where $\theta$ denotes the trainable LoRA parameters. We formulate the multimodal input sequence $x$ by concatenating a task instruction with $\{T, F\}$. 
To distinguish from the discrete class label $y \in \{0, 1, 2\}$ defined in our problem setting, let $Y = \{y_1, y_2, ..., y_N\}$ denote the target response sequence of tokens. This sequence $Y$ contains both the detailed reasoning process and the final textual answer corresponding to the ground-truth label $y$ (i.e., real for $y=0$, cheap-fake for $y=1$, and pure synthesis fake for $y=2$). The target response $Y$ is explicitly formatted as \texttt{<think> [Reasoning across 7 dimensions] </think> <answer> [Textual label for y] </answer>}. The model predicts the \texttt{<think>} trajectory before the final textual \texttt{<answer>} and is optimized via standard autoregressive next-token prediction by minimizing the negative log-likelihood loss,
$\mathcal{L}_\text{SFT} = - \sum_{t=1}^N \log \pi_\theta(y_t | x, Y_{<t})$.

\section{Experiments}
We conduct extensive experiments on the proposed R-T2V framework. In this section, we present the detailed experimental setting and our analysis of the results.

\subsection{Experiment Settings}
\paragraph{PS-FNVD Dataset Construction Details and Evaluation Metrics}
During the PS-FNVD dataset construction phase, we employ GPT-4o \cite{hurst2024gpt} as the prompt generator. To ensure high-quality visual details in the generated T2V prompts while maintaining diversity in scene descriptions, we set the model's temperature to 0.7. 
For the subsequent video synthesis, we utilize Hunyuan Video \cite{kong2024hunyuanvideo} on a server equipped with eight NVIDIA RTX 6000 GPUs Blackwell (96GB VRAM). To balance visual generation quality and computational efficiency, we configure the generation process with $50$ inference steps. Each synthesized news video spans $5$ seconds at a resolution of $544 \times 960$, requiring approximately $5$ minutes of generation time per video.
We partitioned the PS-FNVD dataset into training, validation, and testing sets using an 8:1:1 ratio. As the PS-FNVD dataset features a paired construction where each event (\texttt{video id}) contains both an original news video and its synthetically generated counterpart, a standard random split poses the risk of data leakage. To prevent data leakage, we enforced the dataset split at the \texttt{video id} level. This ensures that the original and generated videos of the same news claim are assigned to the same data subset.
To evaluate the performance of our proposed R-T2V framework, we employ two evaluation metrics, accuracy (Acc.) and macro $F_1$ score.

\begin{table}[!t]
\centering
\caption{Performance compared with the baselines on the PS-FNVD dataset. The best results are highlighted in \textbf{bold}, and the second-best results are \underline{underlined}.}
\label{tab:main_results}
\begin{tabular}{lcc}
\toprule
\multirow{2}{*}{\textbf{Method}} & \multicolumn{2}{c}{\textbf{PS-FNVD}} \\
\cmidrule{2-3}
 & \textbf{Acc.} & \textbf{macro} $F_1$ \\
\midrule
\multicolumn{3}{l}{\textit{\textbf{Zero-shot MLLM Baselines}}} \\
GLM-4V \cite{hong2025glm} & 0.2897 & 0.2730 \\
Qwenvl-2.5-7B-Inst \cite{bai1others} & 0.2929 & 0.2939 \\
Qwenvl-2.5-32B-Inst \cite{bai1others} & 0.3057 & 0.2712 \\
GPT-4o \cite{hurst2024gpt} & 0.3117 & 0.2712 \\
\midrule
\multicolumn{3}{l}{\textit{\textbf{Agent-based Reasoning Baselines}}} \\
ReAct \cite{yao2022react} & 0.6622 & 0.6408 \\
CoRAG \cite{khaliq2024ragar} & 0.6689 & 0.4008 \\
3MFact \cite{niu2025pioneering} & 0.5462 & 0.5478 \\
\midrule
\multicolumn{3}{l}{\textit{\textbf{Supervised Training Baselines}}} \\
FactR1 \cite{zhang2025fact} & 0.6026 & 0.6844 \\
SVFEND \cite{qi2023fakesv} & 0.7180 & 0.7030 \\
ExMRD \cite{hong2025following} & \underline{0.7259} & \underline{0.7154} \\
\midrule
\textbf{R-T2V (Ours)} & \textbf{0.8479} & \textbf{0.8000} \\ 
\bottomrule
\end{tabular}
\end{table}

\paragraph{Baselines and Implementation Details}
We evaluated the performance of our proposed R-T2V framework with ten baselines, four zero-shot MLLM baselines, three agent-based reasoning baselines, and three supervised training paradigm baselines. The four zero-shot MLLM baselines are \textbf{GLM-4V} \cite{hong2025glm}, \textbf{Qwenvl-2.5-7B-Instruct} \cite{bai1others}, \textbf{Qwenvl-2.5-32B-Instruct} \cite{bai1others}, and \textbf{GPT-4o} \cite{hurst2024gpt}. The three agent-based reasoning baselines are \textbf{ReAct} \cite{yao2022react}, \textbf{CoRAG} \cite{khaliq2024ragar}, and \textbf{3MFact} \cite{niu2025pioneering}. The three supervised training paradigm baselines are \textbf{FactR1} \cite{zhang2025fact}, \textbf{SVFEND} \cite{qi2023fakesv}, and \textbf{ExMRD} \cite{hong2025following}.
We first introduce the implementation details of the zero-shot MLLM baselines. For a fair comparison, we apply the identical system prompt used in our proposed R-T2V framework to all MLLM baselines. To ensure reproducibility, the generation temperature of the MLLMs is set to $0$. Consistent with our framework, Qwen-VL \cite{bai1others} and GPT-4o \cite{hurst2024gpt} receive $16$ uniformly sampled frames at the same resolution. The only exception is GLM-4V \cite{hong2025glm}, which is restricted to a $4$-frame input due to the API token constraints.
For the agent-based reasoning baselines, all methods utilize GPT-4o \cite{hurst2024gpt} as the unified backbone model. To ensure a standardized and cost-effective comparison for external knowledge retrieval, we equip all three baselines with the \textit{DuckDuckGo} web search tool rather than their proprietary search setups. Regarding visual inputs, 3MFact \cite{niu2025pioneering} employs its native video sampling strategy, whereas ReAct \cite{yao2022react} processes $16$ frames identical to our visual setting. Since CoRAG \cite{khaliq2024ragar} was originally designed for image-text fake news detection, we adapt it to the video domain by extracting and feeding the first frame of the news video. For the supervised training paradigms, SVFEND \cite{qi2023fakesv} and ExMRD \cite{hong2025following} are trained from scratch and evaluated on our PS-FNVD dataset. 
Following their original hyper-parameter settings,
we make the necessary adaptation by expanding their final classification heads from binary to ternary outputs to align with our T2V-FNVD formulation. Notably, because the original SVFEND \cite{qi2023fakesv} relies on social context features that are naturally absent in our problem setting, we adopt the modality-restricted implementation from \cite{bu2024fakingrecipe} for a fair comparison. Finally, as FactR1 \cite{zhang2025fact} involves a computationally prohibitive LLM fine-tuning process, 
we use the officially released checkpoint for inference on the PS-FNVD testing dataset,
applying the identical system prompt used in our framework to ensure a standardized evaluation.

\begin{table}[!t]
\centering
\caption{Ablation study on the effectiveness of the explicit disentanglement Supervised Fine-Tuning (SFT).}
\label{tab:ablation}
\begin{tabular}{lcc}
\toprule
\multirow{2}{*}{\textbf{Variant}} & \multicolumn{2}{c}{\textbf{PS-FNVD}} \\
\cmidrule{2-3}
 & \textbf{Acc.} & \textbf{macro} $F_1$ \\
\midrule
\textbf{R-T2V (Ours)} & \textbf{0.8479} & \textbf{0.8000} \\
w/o SFT & 0.2929 & 0.2939 \\
\bottomrule
\end{tabular}
\end{table}

\begin{figure*}[t]
\centering 
\includegraphics[scale=1.05]{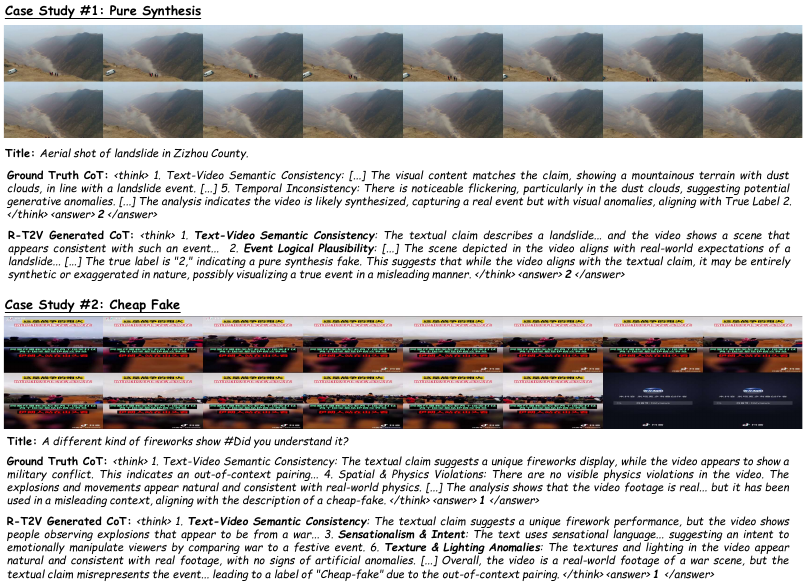}
\caption{Case study: R-T2V correctly disentangles a cheap fake (bottom) from a pure synthesis fake (top).}
\label{case_study}
\end{figure*}

\paragraph{R-T2V Framework Implementation Details}
For our proposed R-T2V framework, we instantiate the base MLLM using Qwen2.5-VL-7B-Instruct \cite{bai1others} and conduct all training and inference utilizing the open-source LLaMA-Factory pipeline \cite{zheng2024llamafactory} on a server equipped with four NVIDIA RTX 6000 Blackwell GPUs (96GB VRAM). To achieve parameter-efficient SFT while preserving the model's pre-trained visual perception capabilities, we freeze both the vision tower and the multimodal projector. Learnable parameters are exclusively updated via Low-Rank Adaptation (LoRA) \cite{hu2022lora} applied to all linear layers within the LLM backbone, configured with a rank of $r=32$, an alpha of $\alpha=64$, and a dropout rate of $0$.
During the training phase, the model is optimized using the AdamW optimizer \cite{loshchilov2017decoupled} with a peak learning rate of $5 \times 10^{-5}$, governed by a cosine learning rate scheduler with $10$ warmup steps. We train the framework for $3$ epochs  using bfloat16 (BF16) precision to optimize memory efficiency. The optimization process operates with a per-device batch size of $1$ and $8$ gradient accumulation steps, capped by a maximum gradient norm of $1.0$. To accommodate the extensive multimodal context alongside the detailed CoT trajectories, the maximum sequence length is truncated at 8{,}192 tokens, and the visual dynamic resolution is constrained to a maximum of 230{,}400 pixels per frame. To balance determinism and diversity in the rationale generation, the decoding parameters are configured with a temperature of $0.7$ and a top-$p$ of $0.7$ during the inference phase on the testing dataset.

\subsection{Experimental Results and Analysis}
Table~\ref{tab:main_results} reports the performance of R-T2V against the baselines, with a per-class breakdown of the same predictions provided in Appendix~\ref{APP:pre-class}; Table~\ref{tab:ablation} reports the SFT ablation. Due to space constraints, four further experiments are deferred to the appendix: cross-benchmark generalization on FakeSV~\cite{qi2023fakesv} and FakeTT~\cite{bu2024fakingrecipe} (Appendix~\ref{APP:Generalization}), explanation quality measured by G-Eval~\cite{liu2023g} (Appendix~\ref{APP:G-Eval}), a modality-reliance ablation (Appendix~\ref{Modality}), and a cross-generator evaluation over five unseen T2V generators (Appendix~\ref{APP:crossT2V}).

\paragraph{Performance Compared with Baselines}
Our proposed R-T2V framework achieves the SOTA performance on the PS-FNVD dataset with an accuracy of $0.8479$ and a macro $F_1$ score of $0.8000$. This substantially outperforms the second-best method (ExMRD) by 12.20 percentage points in accuracy and 8.46 percentage points in macro $F_1$, respectively. Beyond the overall comparison, an in-depth analysis of the baseline paradigms reveals several observations regarding the T2V-FNVD threat.

\textbf{First, parameter scaling alone does not help.} The zero-shot MLLM baselines 
perform poorly across all four models, hovering around $30\%$ accuracy, even below the majority-class baseline of $50.0\%$. Notably, Qwen2.5-VL-7B-Instruct and its drastically larger counterpart, Qwen2.5-VL-32B-Instruct, yield nearly identical, sub-optimal results (e.g., macro $F_1$ of $0.2939$ vs. $0.2712$). This stagnation indicates that blindly scaling up model parameters offers limited robustness against modern generative deceptions. Large MLLMs are heavily victimized by the modality alignment trap; 
their pre-trained cross-modal alignment mechanisms fail to separate the perfect semantic-visual pairings inherent in pure synthesis fake news, leaving their reasoning capacity largely unused.
\textbf{Second, external retrieval plateaus on unverifiable events.} Agent-based reasoning baselines perform moderately better but hit a rigid ceiling (accuracy capping at $0.6689$ for CoRAG). These agents struggle because they rely on web search tools to verify claims. When encountering pure synthesis fake news videos—where fabricators synthesize highly personalized, unrecorded, or emergent events from scratch—search engines return \texttt{not enough evidence}. This unverifiable long-tail problem causes the agents' external reasoning chains to break down, which limits their overall accuracy.
\textbf{Third, supervised baselines remain below 0.73 in accuracy.} While supervised training baselines (FactR1, SVFEND, ExMRD) achieve competitive accuracy ranging from $0.6026$ to $0.7259$, they still trail R-T2V by 12.20 points in accuracy.
These models continue to treat video as a generic semantic container, extracting global representations without explicitly decoupling physical anomalies from the narrative context. Consequently, they struggle to differentiate cheap fakes constructed from existing footage from pure synthesis generated completely from scratch. 
\textbf{Finally, reasoning matters more than model size.} Our R-T2V framework is built upon the Qwen2.5-VL-7B-Instruct backbone, yet it vastly outperforms the zero-shot $32$B version of the exact same model family (by 54.22 percentage points in accuracy). This contrast highlights that the success of R-T2V is not derived from brute-force model capacity, but rather from the explicit disentanglement induced by our Supervised Fine-Tuning (SFT) strategy. By forcing the $7$B model to actively articulate a structured seven-dimensional CoT—jointly evaluating semantic logic and physical generative traces—before generating a prediction, R-T2V successfully changes how the model arrives at its prediction. This indicates that for the T2V-FNVD task, teaching a smaller model \textit{how to reason} via a data-centric reasoning topology is more effective than relying on pre-trained priors alone.

\paragraph{Ablation Study}
To 
assess the contribution of SFT,
we ablate the SFT module, effectively reverting R-T2V to a zero-shot Qwen2.5-VL-7B-Instruct. As shown in Table~\ref{tab:ablation}, removing this reasoning-guided training causes a large performance drop.
Accuracy drops from $0.8479$ to $0.2929$ (a drop of 55.50 percentage points), and Macro $F_1$ similarly 
decreases from 0.8000 to 0.2939.
This degradation indicates that extensive pre-training alone does not confer robustness. Without being explicitly forced to evaluate physical generative traces alongside semantic contexts via our CoT trajectories, the base MLLM 
treats perfectly aligned synthetic visuals as authentic.
Thus, rather than providing a marginal performance gain, our task-specific SFT 
is necessary for the model to perform this reasoning.

\subsection{Case Study}
To intuitively demonstrate the interpretability and explicit disentanglement of our R-T2V framework, we visualize two correctly predicted samples alongside their $16$-frame sequences and reasoning trajectories in Fig.~\ref{case_study}. We utilize a truncated direct quote format to compare the model's output against the ground truth rationale.

As shown in the Cheap Fake sample, the fabricator attempts an out-of-context manipulation by maliciously captioning authentic military conflict footage as a \texttt{fireworks show}. Our R-T2V framework successfully disentangles the multimodal evidence. In the semantic dimensions, it identifies the cross-modal mismatch and explicitly points out the deceptive intent to \texttt{emotionally manipulate viewers}. Concurrently, in the physical dimensions, it confirms that the textures and lighting exhibit \texttt{no signs of artificial anomalies}. This joint evaluation logically leads the model to classify the video as a Cheap Fake (Label 1) composed of authentic pixels. 
Conversely, the Pure Synthesis sample illustrates the modality alignment trap. The generated video depicts a highly realistic landslide that perfectly matches the deceptive textual claim. 
A detector relying on cross-modal consistency checking would find no anomaly here.
However, governed by the structured reasoning topology, R-T2V 
does not rely on surface alignment alone.
The generated CoT explicitly reasons that while the content logically aligns with the text, the video is \texttt{entirely synthetic or exaggerated in nature, possibly visualizing a true event in a misleading manner}. 
By correctly identifying the synthetic origin despite the semantic alignment,
R-T2V successfully classifies it as a pure synthesis fake (Label 2).

\section{Related Work}
In this section, we discuss the related work on T2V-FNVD and the differences from our work. We provide a more comprehensive review of related work in Appendix~\ref{APP:related}.

\paragraph{Text-to-Video Datasets}
Existing datasets fall short of representing the modern T2V FNVD threat. Early datasets primarily focus on text-to-image generation \cite{huang2024miragenews}, lacking the temporal dynamics present in videos. Recently, a wave of large-scale datasets has emerged, which can be categorized into two streams. The first stream focuses on media-published news by modifying text using LLMs or performing non-random entity replacement while keeping the original videos unchanged \cite{wang2024official, wang2025fmnv, zhang2025fact, qi2023fakesv, bu2024fakingrecipe}. These datasets essentially simulate traditional cheap fakes. As discussed earlier, detecting cheap fake news and detecting pure synthesis fake news are two different tasks. The second stream collects massive amounts of purely synthesized AI videos across various generators \cite{ni2026genvidbench, leotta2026synthforensics, veeramachaneni2025leveraging}. However, these datasets are designed purely for cross-source AI video detection and lack the critical fake news context. To the best of our knowledge, our work is the first to introduce a pure T2V FNVD dataset that explicitly models the modality alignment trap.

\paragraph{Fake News Video Detection}
Current methods generally rely on multimodal representation fusion \cite{shang2021multimodal, choi2021using, liu2023covid, qi2023two, qi2023fakesv, bu2024fakingrecipe} or LLM-driven reasoning \cite{hong2025following, niu2025pioneering}. Recently, SOTA frameworks have integrated deep reasoning, Reinforcement Learning (RL) \cite{zhang2025fact}, and active learning \cite{bu2025enhancing}. While effective against cheap fakes, these methods share a hidden assumption: \textit{they treat video merely as a semantic container and rely on internal knowledge (resp.~external knowledge) from LLMs (resp.~search engines) to verify events.} This assumption collapses against T2V synthesis fake news videos due to the modality alignment trap and unverifiable long-tail problem. 

\section{Conclusion}
In this paper, we address the emerging threat of pure synthesis fake news videos by formulating a novel ternary classification task, T2V-FNVD, tailored for the generative AI era. To support this task, we construct PS-FNVD, the first dataset explicitly designed to model the modality alignment trap and semantic-visual degeneration. Furthermore, we propose the R-T2V framework, which achieves explicit disentanglement via SFT on structured reasoning trajectories. By jointly evaluating semantic logic and physical generative traces, R-T2V outperforms existing baselines, demonstrating the effectiveness of a data-centric reasoning approach over standard pre-trained alignment priors.

Despite these contributions, our work has certain limitations that warrant future investigation. First, our framework primarily focuses on the multimodal content itself, neglecting the social propagation dynamics, user engagement, and network structures that often characterize real-world fake news dissemination. Second, regarding the audio modality, the model relies exclusively on text-based audio transcripts, thereby overlooking the nuanced acoustic features and potential AI-generated voice artifacts present in raw audio signals. 
Finally, our ternary taxonomy treats cheap fakes and pure synthesis as distinct categories, but we observe a small number of hybrid videos in FakeSV that splice AI-generated segments into authentic footage. Such mixtures fall between our two fake categories and introduce a degree of label ambiguity for Label 1. 
In future work, we plan to integrate social context graphs, incorporate raw audio analysis, and extend the formulation to segment-level or continuous provenance labeling.

\begin{acks}
This work was supported by the Early Career Scheme (ECS) from the Research Grants Council of HKSAR (HKBU 22202423), the General Research Fund (GRF) from the Research Grants Council of HKSAR (HKBU 12203425), a grant from the Germany/Hong Kong Joint Research Scheme sponsored by the Research Grants Council of HKSAR and the German Academic Exchange Service of Germany (No. G-HKBU208/25), the Initiation Grant for Faculty Niche Research Areas 2023/24 (No. RC-FNRA-IG/23-24/COMM/01), Research Cluster Matching Scheme (No. RCMS/24-25/01) of Hong Kong Baptist University, Guangdong and Hong Kong Universities ``1+1+1'' Joint Research Collaboration Scheme (Project No. 2025A0505000001), National Natural Science Foundation of China (No. 62202402, and No. 61906161), and Startup Grant (Tier 1) for New Academics AY2020/21 of Hong Kong Baptist University.
\end{acks}

\bibliographystyle{ACM-Reference-Format}
\balance
\bibliography{main}

\appendix

\section{In-Depth Methodological Analysis}
This section provides a deeper investigation into the reasoning mechanisms of the proposed R-T2V framework.

\subsection{Per-class Performance of R-T2V on PS-FNVD Testing Dataset}
\label{APP:pre-class}
To provide a clearer understanding of the R-T2V framework's capabilities, we present the per-class performance breakdown in Table~\ref{tab:per_class_metrics}. The results demonstrate that 
R-T2V achieves a recall of 0.9970 and an F1 score of 0.9851 for pure-synthesis fakes (Label 2).
This 
result supports the effectiveness of our explicit disentanglement strategy: by forcing the model to explicitly evaluate physical generative traces via structured reasoning, 
R-T2V is not misled by the modality alignment trap.
The comparatively lower, yet robust, performance on Real News (Label 0) and Cheap Fakes (Label 1) reflects the inherent difficulty of these categories; since both share authentic physical pixels, their distinction relies purely on detecting complex OOC semantic misalignments. Ultimately, 
this breakdown indicates that R-T2V does not merely exploit class priors,
but accurately isolates the distinct physical and semantic anomalies unique to modern generative threats.

\begin{table}[h]
\centering
\caption{Per-class performance breakdown of the proposed R-T2V framework on the PS-FNVD testing dataset.}
\label{tab:per_class_metrics}
\begin{tabular}{lcccc}
\toprule
\textbf{Category (Label)} & \textbf{Precision} & \textbf{Recall} & \textbf{$F_1$-Score} & \textbf{Support} \\
\midrule
Real News (0) & 0.7179 & 0.7044 & 0.7111 & 159 \\
Cheap Fake (1) & 0.7143 & 0.6936 & 0.7038 & 173 \\
Pure Synthesis (2) & \textbf{0.9735} & \textbf{0.9970} & \textbf{0.9851} & 332 \\
\midrule
\textbf{Macro Average} & 0.8019 & 0.7983 & 0.8000 &  \\
\bottomrule
\end{tabular}
\end{table}

\subsection{Modality Reliance}
\label{Modality}
To demonstrate that our proposed R-T2V framework genuinely performs multimodal reasoning and does not exploit the unimodal shortcuts (i.e., making predictions based on isolated textual biases or visual artifacts rather than joint reasoning), we conducted an ablation study on the testing dataset by systematically masking input modalities during inference. We compared the full multimodal R-T2V framework against two restricted variants, i.e., a Text-only setting where all 16 video frames were masked (replaced with blank inputs), and a Vision-Only setting where the textual claim and audio transcript were completely removed. The results are presented in Table~\ref{tab:modality_reliance}.

The results indicate that R-T2V \textbf{relies on both visual and semantic modalities} to achieve its SOTA performance. 
In the \textit{text-only} setting, accuracy decreases from 0.8479 to 0.4729, a drop of 37.50 percentage points, while macro $F_1$ decreases from 0.8000 to 0.3349.
This severe degradation confirms that the paired construction of the PS-FNVD dataset is robust. The textual claims do not contain trivial biases or linguistic leaks that the model can exploit. Without visual evidence, the model
cannot reliably detect the  
generative deceptions. The vision-only setting maintains a high $F_1$ score (0.9800) for pure synthesis fake news videos (Label 2), validating that the model has successfully learned to identify physical generative artifacts purely from pixel data. However, the absence of textual and audio context causes a performance drop in classifying real news (Label 0) from cheap fakes (Label 1), with their $F_1$ scores dropping to 0.6300 and 0.6500, respectively. As cheap fakes are constructed using real-world footage, they lack generative traces. Detecting them strictly requires identifying OOC semantic misalignments between the video and the claim. The full multimodal R-T2V achieves the highest performance across all metrics (Acc: 84.79\%, macro $F_1$: 0.8000). By successfully integrating textual logic with visual artifact detection, the model overcomes the limitations of unimodal approaches, indicating that it actively performs joint reasoning rather than relying on unimodal shortcuts.

\begin{table}[h]
\centering
\caption{Modality reliance test results of R-T2V framework on PS-FNVD testing dataset.}
\label{tab:modality_reliance}
\resizebox{0.48\textwidth}{!}{
\begin{tabular}{llccc}
\toprule
\textbf{Modality Setting} & \textbf{Category (Label)}& \textbf{Acc.} & \textbf{macro} $F_1$ & $F_1$ \\
\midrule
\multirow{3}{*}{\makecell[l]{Text-Only\\(Vision Masked)}} & Real & \multirow{3}{*}{0.4729} & \multirow{3}{*}{0.3349} & 0.1500 \\
& Cheap Fake & & & 0.2300 \\
& Pure Synthesis Fake & & & 0.6200 \\

\multirow{3}{*}{\makecell[l]{Vision-Only\\(Text Masked)}} & Real & \multirow{3}{*}{0.8117} & \multirow{3}{*}{0.7512} & 0.6300 \\
& Cheap Fake & & & 0.6500 \\
& Pure Synthesis Fake & & & 0.9800 \\

\midrule
\multirow{3}{*}{\makecell[l]{\textbf{R-T2V}\\(Full Multimodal)}} & Real & \multirow{3}{*}{\textbf{0.8479}} & \multirow{3}{*}{\textbf{0.8000}} & \textbf{0.7111} \\
& Cheap Fake & & & \textbf{0.7038} \\
& Pure Synthesis Fake & & & \textbf{0.9851} \\
\bottomrule

\end{tabular}
}
\end{table}

\subsection{Quality of G-Eval Explanations}
\label{APP:G-Eval}
\begin{figure}[t]
\centering 
\includegraphics[scale=1.2]{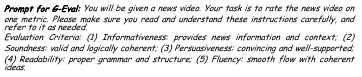}
\caption{Detailed prompt for G-Eval framework.}
\label{fig:G_eval_prompt}
\end{figure}

Following previous studies \cite{hong2025following, wang2024explainable}, we use G-Eval \cite{liu2023g}, an LLM-based reference-free evaluation, to assess explanation quality across five metrics:
(1) Informativeness (I): provides new information and context; (2) Soundness (S): valid and logically coherent; (3) Persuasiveness (P): convincing and well-supported; (4) Readability (R): proper grammar and structure; (5) Fluency (F): smooth flow with coherent ideas. Each metric uses a 5-point Likert scale (1=poorest, 5=best). To ensure reproducibility, we employed GPT-4o as the evaluator model and set the temperature to 0 to minimize randomness and ensure consistent scoring. The evaluation framework and prompt were adapted from the standard G-Eval framework \cite{liu2023g}, instructing the model to rate explanations on a 1-5 Likert scale based on the specific metric definitions. The prompt of the G-Eval framework can be found in Fig.~\ref{fig:G_eval_prompt}. Results on the PS-FNVD dataset (Table~\ref{tab:G_eval}) show that R-T2V achieves the highest scores across all metrics, outperforming even GPT-4o and the much larger Qwen2.5-VL-32B model. Notably, R-T2V achieves a near-perfect soundness score (4.96), far exceeding the Qwen 32B model. This stark contrast demonstrates that explicit SFT on structured CoT trajectories is superior to mere parameter scaling. By learning explicit disentanglement, a compact 7B model is empowered to generate highly logical, persuasive, and fluent forensic explanations that surpass massive SOTA MLLMs.

\begin{table}[t]
\centering
\caption{Comparison of explanation quality for the R-T2V and baseline methods on the PS-FNVD dataset.}
\label{tab:G_eval}
\begin{tabular}{lccccc}
\toprule
\textbf{Methods} & \textbf{I} & \textbf{S} & \textbf{P} & \textbf{R} & \textbf{F} \\
\midrule
GPT-4o \cite{hurst2024gpt} & 4.42 & 4.55 & 4.72 & 4.76 & 4.82 \\
Qwen2.5-VL-7B-Inst (Base) \cite{bai1others} & 4.22 & 4.22 & 4.36 & 4.12 & 4.66 \\
Qwen2.5-VL-32B-Inst \cite{bai1others} & 4.41 & 4.10 & 4.63 & 4.81 & 4.77 \\
\midrule
\textbf{R-T2V} (ours) & 4.57 & 4.96 & 4.82 & 4.84 & 4.91\\
\bottomrule
\end{tabular}
\end{table}

\section{Cross T2V Generator Evaluation}
\label{APP:crossT2V}
To evaluate the cross T2V generator generalization of R-T2V for T2V-FNVD, we conduct an experiment using five unseen T2V generators, they are Seedance (\texttt{seedance-1-5-pro}), Wan (\texttt{wan2.6-t2v}), Sora (\texttt{sora-2}), Veo (\texttt{veo-3-1}), and Kling (\texttt{kling-video}). Specifically, we randomly sampled 62 news videos from the PS-FNVD testing dataset, including 31 Type 1 and 31 Type 2 pure synthesis cases, and regenerated the videos using the same prompts as Hunyuan Video. Three videos failed due to safety filtering, resulting in 307 valid generated videos.
Table~\ref{Tab:Cross} presents the R-T2V results of the cross T2V generator evaluation.
These results show that R-T2V maintains strong cross-generator generalization for PS-FNVD, achieving 93.49\% label 2 recall across five T2V generators. Performance remains high for Seedance, Wan, Veo, and Kling, while Sora is more challenging, with recall dropping to 77.05\%.

\begin{table}[h]
  \centering
  \caption{R-T2V cross T2V generator evaluation results.}
  \begin{tabular}{lcccccc}
    \toprule
    \textbf{Gen.} & \textbf{N} & \multicolumn{2}{c}{\textbf{Label 2}} & \multicolumn{3}{c}{\textbf{Pred.}} \\
    \cmidrule(lr){3-4} \cmidrule(lr){5-7}
    & & \textbf{Corr.} & \textbf{Rec.} & \textbf{0} & \textbf{1} & \textbf{2} \\
    \midrule
    Seedance & 61 & 58 & 95.08\%  & 1 & 2  & 58  \\
    Wan      & 62 & 61 & 98.39\%  & 1 & 0  & 61  \\
    Sora     & 61 & 47 & 77.05\%  & 2 & 12 & 47  \\
    Veo      & 61 & 59 & 96.72\%  & 0 & 2  & 59  \\
    Kling    & 62 & 62 & 100.00\% & 0 & 0  & 62  \\
    \midrule
    \textbf{Overall} & \textbf{307} & \textbf{287} & \textbf{93.49\%} & \textbf{4} & \textbf{16} & \textbf{287} \\
    \bottomrule
  \end{tabular}
  \label{Tab:Cross}
\end{table}

\section{Extended Generalization Evaluation}
\label{APP:Generalization}
To comprehensively evaluate the generalization capabilities of the proposed R-T2V framework, we present the cross-benchmark performance on standard fake news video datasets (FakeSV \cite{qi2023fakesv} and FakeTT \cite{bu2024fakingrecipe}). 

\subsection{Datasets, Baselines, and Metrics}
To evaluate the backward compatibility and cross-domain generalization of our framework, we consider two widely adopted FNVD benchmarks, i.e., FakeSV \cite{qi2023fakesv} and FakeTT \cite{bu2024fakingrecipe}. FakeSV is a large-scale Chinese dataset sourced from prominent short video platforms like \textit{Douyin} and \textit{Kuaishou}, whereas FakeTT provides an English-language context sourced from TikTok.

Consistent with our evaluation on the PS-FNVD dataset, we benchmark the zero-shot generalization performance of our proposed R-T2V framework against ten FNVD baselines spanning three categories: four zero-shot MLLM baselines (GLM-4V \cite{hong2025glm}, Qwen2.5-VL-7B-Instruct \cite{bai1others}, Qwen2.5-VL-32B-Instruct \cite{bai1others}, and GPT-4o \cite{hurst2024gpt}), three agent-based reasoning baselines (ReAct \cite{yao2022react}, CoRAG \cite{khaliq2024ragar}, and 3MFact \cite{niu2025pioneering}), and three in-domain supervised training baselines (FactR1 \cite{zhang2025fact}, SVFEND \cite{qi2023fakesv}, and ExMRD \cite{hong2025following}). For a fair comparison, all zero-shot MLLM baselines are evaluated 
using the same system prompt as our R-T2V framework, while the remaining baselines are reported based on their official implementations. 
We clarify the dataset splitting strategy to address potential data leakage.
Although the validation and test sets of our PS-FNVD dataset were independently shuffled and differ from the official splits of FakeSV, we strictly enforced a unidirectional exclusion protocol during dataset construction. Specifically, original videos from the official FakeSV validation and test sets were completely excluded from the PS-FNVD training phase. Consequently, the R-T2V framework was never exposed to these specific events or their multimodal content during training, ensuring that our cross-benchmark evaluations on the FakeSV and FakeTT testing datasets are conducted on genuinely held-out, unseen data distributions.

For the evaluation metrics, we utilize Acc. and macro $F_1$ score. As traditional benchmarks formulate FNVD as a binary classification task (Real vs. Fake), we adopt a \textit{binary detection} metric for fair comparison. Specifically, any model prediction indicating deception, whether Label 1 (Cheap Fake) or Label 2 (Pure Synthesis Fake), is aggregated into a unified \textit{Fake} category.

\subsection{Cross-Benchmark Generalization Performance Analysis}
Table~\ref{tab:cross_benchmark} presents the comprehensive cross-benchmark generalization performance.

\paragraph{Backward Compatibility on FakeSV} Despite having zero exposure to the FakeSV test distribution during training, R-T2V demonstrates strong cross-domain generalization, achieving a binary Acc. of $0.7536$ and a macro $F_1$ of $0.7514$. Crucially, we observe a massive performance leap compared to its non-fine-tuned foundation model (Qwen2.5-VL-7B-Instruct), which only scores an $F_1$ of $0.4752$. Through our SFT on the PS-FNVD dataset, R-T2V overcomes the foundation model's naive cross-modal alignment bias, vastly outperforming even GPT-4o ($F_1$ 0.6117) and the 32B parameter variant of Qwen ($F_1$ $0.6429$). Furthermore, our zero-shot R-T2V exhibits backward compatibility that rivals heavily optimized, fully in-domain supervised baselines (e.g., FactR1 at $0.7478$), suggesting that explicit reasoning over physical traces transfers to detecting traditional cheap fakes.

\paragraph{Validation of Explicit Disentanglement} Beyond binary classification, we rigorously audit the model's ternary predictions on FakeSV to verify its disentanglement capability. Since FakeSV contains no pure synthesis videos, any prediction of Label 2 constitutes a failure to disentangle generative artifacts from traditional editing. Impressively, out of the hundreds of test samples, R-T2V incorrectly predicted Label 2 in only 6 cases.
This near-perfect strict accuracy on traditional fakes provides evidence that our seven-dimensional reasoning topology successfully isolates physical generative traces, 
indicating that the model does not frequently overpredict the \textit{pure-synthesis} label on FakeSV.

\begin{table}[h]
    \centering
    \caption{Cross-benchmark generalization performance on FakeSV and FakeTT datasets. All evaluations adopt the Binary Detection metric. Supervised baselines are trained in-domain, whereas MLLM, Agent-based, and our R-T2V frameworks are evaluated in a zero-shot, cross-domain setting.}
    \label{tab:cross_benchmark}
    \resizebox{\linewidth}{!}{
    \begin{tabular}{llcccc}
        \toprule
        \multirow{2}{*}{\textbf{Paradigm}} & \multirow{2}{*}{\textbf{Method}} & \multicolumn{2}{c}{\textbf{FakeSV (Binary)}} & \multicolumn{2}{c}{\textbf{FakeTT (Binary)}} \\
        \cmidrule(lr){3-4} \cmidrule(lr){5-6}
        & & Acc. & macro $F_1$ & Acc. & macro $F_1$ \\
        \midrule
        \multirow{4}{*}{\shortstack[l]{Zero-shot\\MLLM}} 
        & GLM-4V & 0.4097 & 0.3040 & 0.5209 & 0.3647 \\
        & Qwen2.5-VL-7B-Inst (Base) & 0.4969 & 0.4752 & 0.5856 & 0.5278 \\
        & Qwen2.5-VL-32B-Inst & 0.6471 & 0.6429 & 0.6730 & 0.6631 \\
        & GPT-4o & 0.6227 & 0.6117 & 0.6730 & 0.6580 \\
        \midrule
        \multirow{3}{*}{\shortstack[l]{Agent-based\\Reasoning}} 
        & ReAct & 0.6084 & 0.6016 & 0.6519 & 0.6643 \\
        & CoRAG & 0.5918 & 0.5833 & 0.6720 & 0.6889 \\
        & 3MFact & 0.8122 & 0.8359 & 0.7661 & 0.7920 \\
        \midrule
        \multirow{3}{*}{\shortstack[l]{Supervised\\(In-domain)}} 
        & FactR1 & 0.7662 & 0.7478 & 0.7444 & 0.7270 \\
        & SVFEND & 0.8086 & 0.8055 & 0.7714 & 0.7563 \\
        & ExMRD & 0.8490 & 0.8512 & 0.8428 & 0.8315 \\
        \midrule
        \textbf{Ours} & \textbf{R-T2V (Zero-shot)} & \textbf{0.7536} & \textbf{0.7514} & 0.4905 & 0.4851 \\
        \bottomrule
    \end{tabular}
    }
\end{table}

\paragraph{Error Analysis on Extreme Domain Shifts (FakeTT)} While R-T2V excels on FakeSV, its performance encounters a bottleneck on the FakeTT dataset (Binary $F_1$ $0.4851$), ostensibly underperforming its non-fine-tuned base model ($F_1$ $0.5278$). However, a deeper diagnostic reveals that this constraint is caused by a severe visual domain shift and statistical class imbalance, rather than a degradation of forensic capability. FakeTT comprises predominantly short-form TikTok videos characterized by heavy platform-specific aesthetics (e.g., erratic filters, extreme jump cuts, and sensational text overlays) \cite{sun2020content, kaye2021co}. When R-T2V evaluates the real TikTok news videos via its seven-dimensional reasoning topology, these stylistic edits frequently trigger the \textit{Texture Anomalies} and \textit{Sensationalism} detectors, leading to a high False Positive rate (i.e., over-flagging real TikTok as manipulated). Conversely, while the base Qwen-7B model 
achieves higher aggregate accuracy and macro $F_1$, this is a statistical artifact masking a forensic failure. Due to the dataset's extreme imbalance (175 Real vs. 88 Fake), the base model exhibits extreme inertia, defaulting to real predictions and missing nearly all actual fake news (Fake Recall: $0.3520$). In contrast, R-T2V achieves a higher Fake Recall ($0.8860$), demonstrating strong zero-shot sensitivity to deception, albeit at the cost of precision in heavily stylized, out-of-distribution environments.

\begin{figure*}[t]
\centering 
\includegraphics[scale=1]{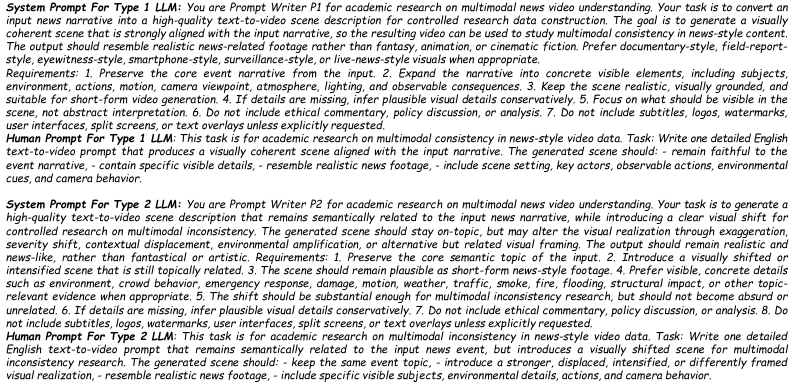}
\caption{Detailed prompts for T2V generation.}
\label{t2v_prompt}
\end{figure*}

\begin{figure*}[t]
\centering 
\includegraphics[scale=1]{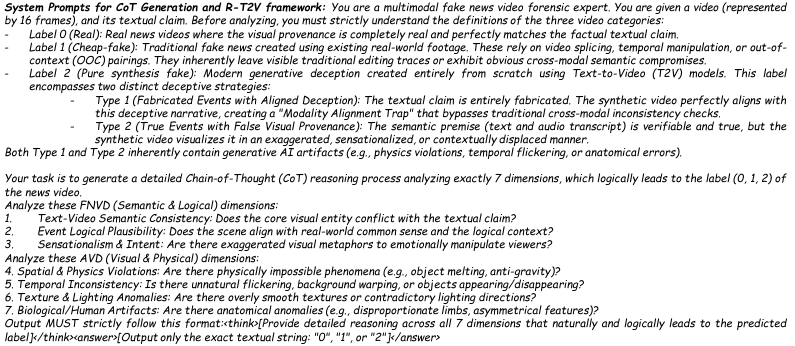}
\caption{System Prompts for CoT generation and R-T2V Framework.}
\label{cot_rt2v_prompt}
\end{figure*}

\section{Dataset Construction and Implementation Details}
\label{APP:Dataset}
To support reproducibility and provide deeper insights into the PS-FNVD dataset construction, this section details the specific configurations utilized. We provide the detailed prompts and safety bypass strategies for T2V generation, along with the complete system prompt driving the R-T2V framework. Furthermore, we outline the ethical considerations, access safeguards established for the dataset, and code availability.

\subsection{Detailed Prompts for T2V Generation and Safety Bypass Strategies}
Modern LLMs and T2V generation models are equipped with strict safety alignment and moderation mechanisms. Directly inputting fake news texts involving disasters, conflicts, or sensitive events often triggers these filters, leading to generation failures. To successfully construct a dataset for defensive research without violating the underlying safety guidelines of these models, we strategically incorporated the following bypass techniques into our prompt design:
\textbf{Academic Framing}: We explicitly start all System and Human Prompts with a declaration that the task is for \texttt{academic research on multimodal news video understanding}. This contextual framing effectively signals the legitimate intent of the request, reducing the likelihood of the model misinterpreting the generation as malicious forgery.
\textbf{Objective Visual Focus}: By strictly prohibiting subjective ethical judgments or policy analysis (e.g., \texttt{Do not include ethical commentary, policy discussion, or analysis}), we force the LLM to focus entirely on purely physical visual elements (e.g., lighting, environment, crowd actions, weather). This helps bypass scrutiny related to political or ideological sensitivities.
\textbf{Moderated Exaggeration and Entity Generalization}: When generating exaggerated scenes for Type 2 (false visual provenance), we guide the model to emphasize environmental and structural visual impacts (e.g., \texttt{smoke, fire, flooding, structural impact}) rather than generating extreme gore or violence targeting specific real-world individuals. This keeps the generation success rate of T2V models high while fulfilling the necessary requirements for multimodal inconsistency research.

The core objective of the Type 1 Prompt Writer is to transform an input fabricated news narrative into a highly realistic, visually coherent scene description that is strongly aligned with the text. This is designed to create a \textit{Modality Alignment Trap} within the dataset, challenging models to detect fabrications where the visual and textual modalities perfectly corroborate each other. The objective of the Type 2 Prompt Writer is to take a real news event and generate a scene description that remains semantically related but introduces a clear visual shift or intensification. This simulates the real-world behavior of malicious actors attaching false or exaggerated visual evidence to true events, providing controlled samples for multimodal inconsistency research. The detailed prompts for Type 1 and Type 2 Prompt Writers are presented in Fig.~\ref{t2v_prompt}.

\subsection{System Prompts for CoT generation and R-T2V Framework} 
We designed a structured system prompt that enforces explicit disentanglement during both the generation of the CoT reasoning corpus and the subsequent SFT phase. The detailed prompts are presented in Fig.~\ref{cot_rt2v_prompt}. The first section of our prompt explicitly calibrates the model's understanding of the T2V-FNVD taxonomy. By defining the boundaries of \textit{Label 1 (cheap fake)} and \textit{Label 2 (pure synthesis fake)}, we force the model to acknowledge the specific mechanisms of modern deception. Crucially, we embed the definitions of \textit{Type 1 (fabricated events with aligned deception)} and \textit{Type 2 (True events with false visual provenance)} directly into the prompt. This a priori knowledge prevents the model from defaulting to \textit{Real} when encountering fabricated events with perfectly aligned visuals, and stops it from ignoring generative artifacts when the textual premise is true.
Traditional FNVD relies heavily on cross-modal semantic inconsistencies checking, while AVD strictly targets visual artifacts. Our prompt unifies these disparate paradigms by explicitly demanding an analysis across 7 predefined dimensions. \textbf{FNVD Semantic \& Logical Dimensions:} Designed to detect the narrative compromises and OOC pairings inherent in traditional cheap fakes. \textbf{AVD Visual \& Physical Dimensions:} Inspired by recent SOTA AVD methods \cite{li2025skyra, park2025vidguard}, we designed the prompt to expose the pure synthesis fake generated completely from scratch. Dimensions like \textit{Texture \& Lighting Anomalies} force the model to scrutinize low-level generative traces.

\subsection{Ethical Statement}

The emergence of pure synthesis fake news videos poses a severe threat to public information integrity. Our research necessitated generating realistic deceptive content for defensive purposes. We acknowledge that constructing PS-FNVD involved bypassing commercial safety filters of LLMs and T2V generators, conducted strictly under the paradigm of \textit{defensive research}.
To mitigate potential harms, our prompt rewriting strategies (e.g., moderated entity generalization) deliberately avoided producing explicitly illegal, excessively violent, or severe hate speech content. Generated videos simulate realistic misinformation contexts (political events, natural disasters) strictly for forensic evaluation, without crossing into severely toxic materials.

\section{Comprehensive Related Work}
\label{APP:related}
This section provides an extended review of the related literature, situating the T2V-FNVD formulation within the broader context of recent advancements.

\subsection{Text-to-Video Datasets}
Existing datasets fall short of representing the modern T2V fake news video threat. Early datasets primarily focus on text-to-image generation \cite{huang2024miragenews}, lacking the temporal dynamics present in videos. Recently, a wave of large-scale datasets has emerged, which can be categorized into two streams. The first stream focuses on media-published news by modifying text using LLMs or performing non-random entity replacement while keeping the original videos unchanged \cite{wang2024official, wang2025fmnv, zhang2025fact, qi2023fakesv, bu2024fakingrecipe}. 
FakeSV \cite{qi2023fakesv} takes a comprehensive multimodal and social context perspective by compiling the largest collection of Chinese short videos alongside their corresponding news content, user comments, and publisher profiles. FakeTT \cite{bu2024fakingrecipe} is constructed by retrieving TikTok videos linked to Snopes fact-checking reports. The Official-NV dataset \cite{wang2024official} utilizes LLMs to augment the data scale by rewriting textual news title (without modifying the video frames), across dimensions like position, quantity, action, and object to generate news texts with either the same or opposite meanings. Based on Official-NV dataset \cite{wang2024official}, FMNV \cite{wang2025fmnv} incorporates the audio transcripts and introduces a more detailed classification for OOC cheap fakes. FakeVV dataset \cite{zhang2025fact} is constructed by employing LLMs to perform non-random entity replacements (i.e., modifying text entities like person and locations) in real news titles without altering the video frames to generate OOC cheap fake news videos.
These datasets essentially simulate traditional cheap fakes. As discussed earlier, cheap fake news detection and pure synthesis fake news detection are two different tasks. The second stream collects massive amounts of purely synthesized AI videos across various generators \cite{ni2026genvidbench, leotta2026synthforensics, veeramachaneni2025leveraging}. However, these datasets are designed purely for cross-source AI video detection and lack the critical fake news context. To the best of our knowledge, our work is the first to introduce a pure T2V fake news video dataset that explicitly models the semantic-visual alignment trap.

\subsection{Fake News Video Detection}
Current methods generally rely on multimodal representation fusion \cite{shang2021multimodal, choi2021using, liu2023covid, qi2023two, qi2023fakesv, bu2024fakingrecipe, tian2026exposing} or LLM-driven reasoning \cite{hong2025following, niu2025pioneering}. 
\citeauthor{shang2021multimodal} \cite{shang2021multimodal} propose TikTec to detect misleading Covid-19 short videos by utilizing textual captions. \citeauthor{liu2023covid} \cite{liu2023covid} propose TwtrDetective, which utilizes cross-modal consistency checking to detect token-level tampering and generate explanations for Covid-19 short videos. \citeauthor{choi2021using} \cite{choi2021using} propose a topic-agnostic FNVD method that leverages topic modeling to identify stance discrepancies between video metadata and user comments. \citeauthor{qi2023fakesv} \cite{qi2023fakesv} propose SVFEND that leverages cross-modal correlations for feature selection and integrates social context information to improve FNVD. 
\citeauthor{bu2024fakingrecipe} \cite{bu2024fakingrecipe} take the perspective from the fake news video creation process and propose FakingRecipe. FakingRecipe identifies manipulated short videos by analyzing the sentimental, semantic, spatial, and temporal traits of material selection and editing.
\citeauthor{hong2025following} \cite{hong2025following} introduce ExMRD that utilizes LLM for fake news video data augmentation. \citeauthor{niu2025pioneering} \cite{niu2025pioneering} introduce 3MFact that iteratively gathers and synthesizes online evidence to progressively generate veracity labels of news videos. \citeauthor{tian2026exposing} \cite{tian2026exposing} discovered cases where modalities appear superficially consistent but contain factual errors, yet their methods still rely heavily on cross-modal inconsistency for detection. When facing the modality alignment trap, purely synthetic videos (Pure Synthesis) generated from scratch by T2V models are perfectly aligned semantically, causing purely consistency-centric models to fail.
Recently, SOTA frameworks have integrated deep reasoning, Reinforcement Learning (RL) \cite{zhang2025fact}, and active learning \cite{bu2025enhancing}.
While effective against cheap fakes, these methods share a hidden assumption: \textit{they treat video merely as a semantic container and rely on internal knowledge (resp.~external knowledge) from LLMs (resp.~search engines) to verify events.} This assumption collapses against T2V synthesis fake news videos due to the modality trap and unverifiable long-tail problem.

\clearpage

\end{document}